\documentclass{llncs}

\usepackage[english]{babel}

\usepackage[letterpaper,top=2cm,bottom=2cm,left=3cm,right=3cm,marginparwidth=1.75cm]{geometry}

\usepackage[T1]{fontenc}
\usepackage{amsmath}
\usepackage{amssymb}
\usepackage{graphicx}
\usepackage[colorlinks=true, allcolors=blue]{hyperref}

\usepackage{multirow}
\usepackage{caption,subcaption}
\usepackage{tabularx}
\usepackage{booktabs}
\usepackage{bbding}

\newcommand{\PreserveBackslash}[1]{\let\temp=\\#1\let\\=\temp}
\newcolumntype{C}[1]{>{\PreserveBackslash\centering}p{#1}}
\newcolumntype{R}[1]{>{\PreserveBackslash\raggedleft}p{#1}}
\newcolumntype{L}[1]{>{\PreserveBackslash\raggedright}p{#1}}
\newcolumntype{Y}{>{\centering\arraybackslash}X}

\title{Unifying Runtime Monitoring Approaches for Safety-Critical Machine Learning: Application to Vision-Based Landing}

\author{
    Mathieu Dario  \inst{1,2,3} \Envelope \and 
    Florent Chenevier  \inst{3} \and 
    Kévin Delmas  \inst{4} \and 
    Joris Guerin  \inst{5,6} \and 
    Jérémie Guiochet \inst{1,2}
}

\institute{
    LAAS-CNRS, Toulouse, France \\ 
    \email{\{firstname.lastname\}@laas.fr} 
    \and
    University of Toulouse, France
    \and
    Thales, Toulouse, France 
    \and
    ONERA, Toulouse, France
    \and 
    Espace-Dev, IRD, Université de Montpellier, Montpellier, France
    \and
    Universidade Federal do Rio Grande do Norte, Natal-RN, Brazil
}

\begin{document}
\maketitle

\begin{abstract}
Runtime monitoring is essential to ensure the safety of ML applications in safety-critical domains. However, current research is fragmented, with independent methods emerging from different communities. In this paper, we propose a unified framework categorising runtime monitoring approaches into three distinct types: Operational Design Domain (ODD) monitoring, which ensures compliance with expected operating conditions; Out-of-Distribution (OOD) monitoring, which rejects inputs that deviate from the training data; and Out-of-Model-Scope (OMS) monitoring, which detects anomalous model behaviour based its internal states or outputs. We demonstrate the benefits of this categorization with a dedicated experiment on an aeronautical safety-critical application: runway detection during landing. This framework facilitates design of monitoring activities, with complementary categories of monitors, and enables evaluation and comparison of different monitors using common, safety-oriented metrics.

\end{abstract}

\section{Introduction} \label{sec:1-introduction}

Recent advances in Machine Learning (ML) technologies have enabled their deployment across a range of domains, including safety-critical applications such as autonomous driving, robotic surgery, and aviation, where data-driven techniques promise enhanced automation and decision support. In particular, deep neural networks (DNNs) underpin state-of-the-art perception methods for tasks like image classification and object detection, interpreting complex sensor signals to provide state estimates used in decision and control. However, despite impressive achievements, DNNs inherently lack robustness and reliability~\cite{amodei2016concrete}, with performance degrading under unforeseen or adverse inputs~\cite{goodfellow2014explaining}, limited explainability~\cite{montavon2018methods}, and often excessive confidence when mispredicting~\cite{nguyen2015deep}. To address these challenges, safety standards for ML integration in critical systems are evolving, exemplified by initiatives from entities such as the European Aviation Safety Agency (EASA)~\cite{EASA_AI_Roadmap_1_2020}. Runtime safety monitors, derived from the dependability community, have emerged as promising solutions to improve the reliability of ML functions. They are fault-tolerance mechanisms, continuously monitoring the behaviour, inputs, and outputs of ML components. Through early anomaly or failure detection, these monitors can trigger mitigation at runtime, providing an additional safeguard beyond traditional offline verification.

While various methods have been developed to monitor the safety of ML components, they often come from distinct research communities. For instance, the dependability community and the ML community both provided approaches that rely on different concepts within different scopes (global system operational domain vs. ML-specific detection of anomalies). This fragmentation results in partial coverage of the safety risks, each methodology focusing on different aspects of the problem. The limitations arising from this lack of unified perspective are further examined in Section~\ref{sec:2-related-work}.

\begin{figure}[t]
    \centering
    \begin{subfigure}{0.64\textwidth}
        \centering 
        \includegraphics[width=0.95\linewidth]{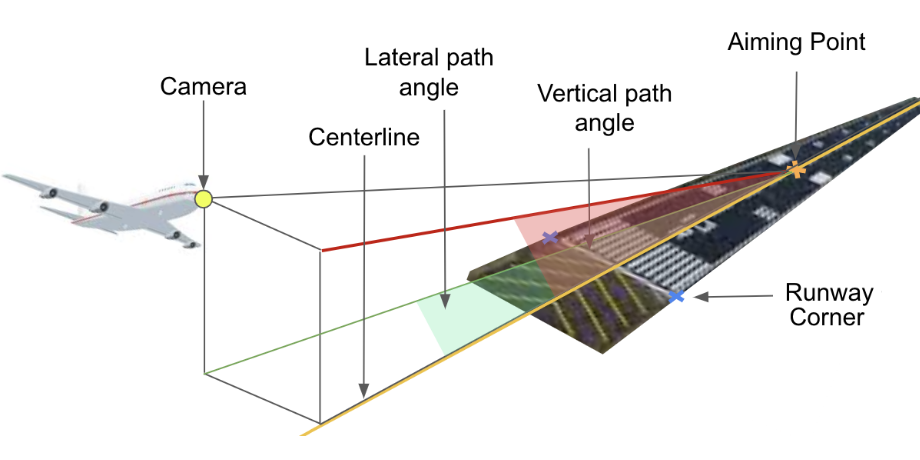}
    \end{subfigure} 
    \hfill
    \begin{subfigure}{0.32\textwidth}
        \centering
        \includegraphics[width=0.95\linewidth]{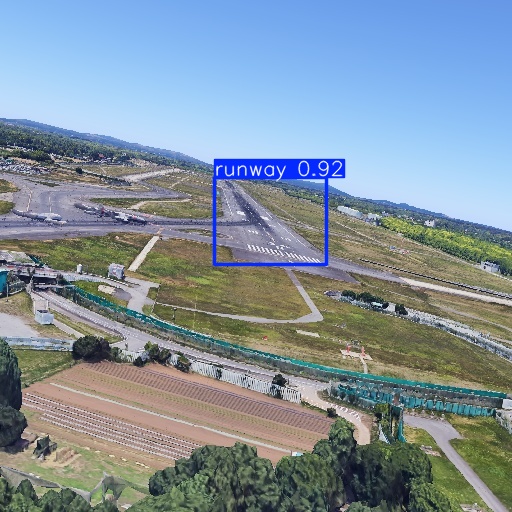}
    \end{subfigure}
    \caption{
        Illustration of the Vision-Based Runway Detection (VBRD) task.
    }
    \label{fig:vbl-use-case}
\end{figure}

In this paper, we clarify the definitions of the key monitoring notions used across the different research communities, establishing a common understanding and terminology. We introduce a unified, multi-level framework called SwMF, to organise monitoring approaches according to the risks they address in the ML input space. Finally, we illustrate our framework through an experimental study on an aviation safety-critical application, namely Vision-Based Landing (VBL)~\cite{cappi2024how}. We particularly focus on the sub-task of detecting the runway in an image taken during landing, as depicted in Figure~\ref{fig:vbl-use-case}.

The rest of the paper is organized as follows. Section \ref{sec:2-related-work} presents a review of current approaches related to ML safety monitoring. Section \ref{sec:3-my-theory} introduces our proposed formalism (SwMF), by redefining data domains and data threats to derive a threat-based unified classification of monitoring activities. Section \ref{sec:4-my-experiment} details our experimental study, illustrating how our framework can be implemented on a concrete use case. Finally, Section \ref{sec:5-conclusion} presents our conclusions and perspectives.

\section{Related Work} \label{sec:2-related-work}

In this section, we present previous contributions in the domain for runtime safety monitoring of ML functions. We first give an overview of the sector of traditional safety monitoring before presenting its adaptation to the specific case of ML models.

\subsection{From Traditional Safety Monitoring to ML Monitoring}

In the dependability community~\cite{avizienis2004basic}, safety monitors (SMs) are fault-tolerance mechanisms, observing system inputs, states, and outputs as well as environment states to ensure continued safe operation. They verify compliance with predefined \textit{safety properties}, established via rigorous system-level safety analyses and formalized by domain experts. They are positioned at critical decision points within the system architecture, ensuring that only safe states are propagated downstream. This approach is well-established and effective when system and environment states can be measured and hazards can be identified and formalized.

The integration of ML models in perceptions tasks, such as DNNs that estimate system states from sensor inputs (images, videos), brings the need for ML-specific monitors. Indeed, applying SMs to ML models presents significant challenges. Traditional SMs can be used in a black-box manner to monitor systems~\cite{machin2016smof}, including ML systems, by leveraging redundant sources of observation to infer system or environment states and verify corresponding safety rules. However, this becomes particularly difficult when dealing with complex ML-based tasks such as computer vision, where no straightforward alternative exists to DNNs. Additionally, developing interpretable safety properties at the ML level is also problematic. Expressing rules or assertions based on internal model features, such as neuron values or layer activations, remains opaque and uninformative from a system safety perspective. To address these limitations, data-driven monitors have emerged. They learn statistical patterns observed during model training to correlate inputs, outputs, or internal features with possible failure risks. Despite recent advances, substantial unresolved questions remain, as pinpointed by Ferreira et al.~\cite{ferreira2025safety}, such as identifying relevant threats to address, translating system safety properties to ML-level criteria, designing efficient mechanisms for both detecting and mitigating ML failures, and finally evaluating monitoring impact on system safety. Diverse communities have addressed the question of ML safety monitoring, in rather independent ways. System-level approaches, like SMs, were developed by the dependability community, for instance. ML-level approaches were developed in the ML community. In the next sections, we present the main initiatives for monitoring the ML safety at runtime. We categorize these approaches into three categories, that are roughly the monitoring of the system's intended usage conditions (Section~\ref{sec:2-sota-odd}), the detection of statistically anomalous input data (Section~\ref{sec:2-sota-ood}), and the detection of input data that lead to system failures (Section~\ref{sec:2-sota-oms}).

\subsection{Monitoring the Operational Design Domain (ODD)} \label{sec:2-sota-odd}

The concept of Operational Design Domain (ODD) emerged first in the automotive safety domain as the set of conditions under which an automated driving system is intended to safely operate~\cite{saeinternational2021taxonomy}. These conditions typically cover a broad range of constraints, including environmental factors, geographic restrictions, and temporal or situational boundaries. In aviation, the EASA has adapted the ODD concept in the context of ML integration in broader systems, defining it as ``\textit{the operating conditions under which a given AI/ML constituent is specifically designed to function as intended, including but not limited to environmental, geographical, and/or time-of-day restrictions}''~\cite{EASA_AI_Concept_Paper_2_2024}. Monitoring the ODD does not focus on safety conditions like traditional safety monitoring. It  concerns the definition of boundaries of the system's functionality, with the rationale that, out of these boundaries, the system's functionality performances might decrease (without necessarily leading to a safety issue).

However, defining the ODD for ML-based components remains a complex task. Indeed, the ODD concept itself is not well-defined, spanning from the \textit{ideally} expected operating conditions to the operational domain the engineers can truly specify. Moreover, in most cases, it is because of the absence of a well specifiable operational domain that ML models are used: they can learn from examples when full system specifications are not possible. Still, for system-level functionalities, it is possible to specify operating conditions in terms of constraints on system variables such as position, velocity, external weather, etc. However, ML components, especially those handling high-dimensional sensory data, raise the question of whether the ODD should be defined at the system integration level or more directly at the ML-level in terms of input characteristics. Existing approaches illustrate this ambiguity. Some works monitor the ODD at the system level, ensuring at runtime that system variables like vehicle state or environment satisfy the requirements established during design~\cite{torens2025runtime}. On the contrary, Kaakai et al.~\cite{kaakai2023datacentric} proposes a data-centric approach for characterizing the ODD, bringing system safety concerns to the ML level. Finally, Cappi et al.~\cite{cappi2024how}, bridges the gap between system and ML levels, refining a system ODD specification down to ML level requirements in terms of image characteristics, illustrating the process on a Vision-based Landing (VBL) system. They verify the relevance and representativeness of the LARD (Landing Approach Runway Detection)~\cite{ducoffe2023lard} dataset to the specified ODD. Thus, in practice, ODD monitoring ranges from approaches centred on high-level operational parameters to methods that define the ODD at the ML-level, through data ranges and distributions. This variety highlights a key challenge: ODD boundaries at the system level are not always directly aligned with data-domain boundaries relevant for ML elements. 

\subsection{Monitoring Out-of-Distribution (OOD) Data} \label{sec:2-sota-ood}

Out-of-Distribution (OOD) monitoring has emerged, in the ML community, as a key ML-centric approach to validate the reliability of ML models at runtime. The core idea is to identify inputs that substantially differ from those encountered during ML training. While conceptually appealing, the definition of OOD remains nuanced, spanning a continuum of interpretations. In the ML safety community, OOD data are generally defined as instances not drawn from the training distribution $\mathcal{D}_{ID}$, also called ``in-distribution''. The OOD detection task consists of detecting and rejecting such inputs. The rationale behind OOD monitoring is that DNNs tend to perform well on data drawn from the same distribution than training data, but show degraded performance on data from outside this distribution. Additionally, the practical need for robust OOD monitoring is underscored by several safety authorities, like the EASA~\cite{EASA_AI_Concept_Paper_2_2024}. In practice, OOD may refer to a wide range of situations, from statistical shifts in raw input data to fundamental changes in the input data content. The literature commonly distinguishes two main types of distribution shifts: \textit{covariate} and \textit{semantic} shifts~\cite{ferreira2025safety}. On the one hand, covariate shifts refer to variations in input features (such as those caused by changing weather, lighting, or sensor characteristics) impacting the distribution of data form but not the underlying semantics~\cite{hendrycks2018benchmarking}. On the other hand, semantic shifts concern more fundamental distributional changes in scene content (such as novel classes of objects, novel attributes of known classes, or atypical objects interactions).

Detection strategies cover a broad range of approaches. Some analyse the incoming inputs, flagging distribution differences from model's training datasets. These correspond to traditional signal processing approaches that look for statistical patterns in the input before it enters the model~\cite{ndong2011signal}. More recent techniques rely on deep autoencoders to identify anomalous images, either computing reconstruction errors or performing outlier detection in latent spaces~\cite{cai2020real,denouden2018improving}. Other works look at intermediary features of the ML model itself to infer anomalous behaviour. These correspond to neuron activation patterns techniques such as box-abstraction monitors~\cite{he2024boxbased,henzinger2020outside}. Distance-based and density-based detectors using Mahalanobis distance scores~\cite{lee2018simple}, k-Nearest Neighbors~\cite{sun2022outofdistribution}, or Virtual logits Matching~\cite{wang2022vim} further expand this spectrum. Finally, some approaches leverage the ML outputs. A pioneer work~\cite{hendrycks2017baseline} proposes a baseline, applying softmax confidence thresholds. Other output-based scoring methods include ODIN~\cite{hsu2020generalized}, ReAct~\cite{sun2021react}, and Energy-based detector~\cite{liu2020energybased}. Most of the detectors described above were designed for monitoring image classification models, and so far, very few techniques are proposed for the more complex task of object detection. Still, Du et al.~\cite{du2022vos} proposes VOS, a pioneer OOD detector for object detection, followed by later OOD detector SIREN~\cite{du2022siren}. Also, Wilson et al.~\cite{wilson2023safe} proposes a detector based on SAFE vectors extracted from the object detection model. Some approaches used in image classification were also adapted (to some extent) to object detection, mostly to derive baselines to compare the methods with. Finally, comprehensive surveys~\cite{yang2024generalized} and toolkits that aggregate OOD techniques, such as \texttt{OODEEL}~\cite{oodeel}, \texttt{pytorch-ood}~\cite{kirchheim2022pytorchood}, and \texttt{OpenOOD}~\cite{zhang2023openood} are available. 

\subsection{Monitoring Out-of-Model-Scope (OMS) Data} \label{sec:2-sota-oms}

Out-of-Model-Scope (OMS) monitoring provides an important refinement over the traditional OOD paradigm, a distinction recently articulated by Guerin et al.~\cite{guerin2023outofdistribution}. Unlike OOD monitoring, which is defined regarding deviations from training data distribution, OMS monitoring is intrinsically tied to a specific ML function $f$ and its reliable operating region. Guerin et al. introduces the notion of model scope $\mathcal{D}_f$, representing the subset of inputs on which the model is acceptably accurate or ``safe''. OMS samples are defined as data points lying outside this zone, i.e., inputs for which the model cannot guarantee appropriate performances, regardless of their proximity to the training distribution. This distinction is significant. As shown by Guerin et al., OOD detection provides, at best, a rough proxy for actual model errors: a model may generalise successfully to certain OOD samples, and conversely, can fail even on in-distribution samples. Recognizing these nuances, OMS monitoring targets the more direct problem of detecting and rejecting inputs that will cause the supervised ML model to fail.

Several monitoring approaches, often initially presented in the context of OOD, address the OMS challenge either explicitly or through their evaluation strategies. For instance, Wang et al.~\cite{wang2020dissector} proposes Dissector to detect OMS inputs by analysing neuron-level activation patterns. Similarly, Cheng et al.~\cite{cheng2019runtime} proposes a pattern-based monitor evaluated specifically in terms of their capacity to detect ML errors, regardless of strict OOD/ID designation. Moreover, Hashemi et al.~\cite{hashemi2024gaussianbased} proposes hybrid monitors combining Gaussian-based and box-abstraction methods, that are designed and evaluated under the OMS paradigm. Methods for monitoring ML model performance are closely related to OMS detection. For example, Rahman et al.~\cite{rahman2019did} proposes a False Negative Detector (FND) for traffic sign detection. Subsequent research develops performance runtime monitors that observe ML models’ internal features during deployment~\cite{rahman2021perframe}. Additionally, Chen et al.~\cite{chen2021monitoring} introduces a runtime monitor that analyses the outputs of object detectors to identify abnormal label attributes (e.g.,  size, location) and temporally incoherent predictions (e.g., label flips, object disappearance).

It is important to emphasize that OMS and OOD techniques often overlap methodologically, with the primary distinction residing in their intended objective and means of evaluation. While many detectors developed under the OOD framework can be employed for OMS detection, the OMS perspective specifically evaluates monitors based on their ability to anticipate errors in the deployed model. It is, thus, necessary to distinguish the two paradigms and to clarify and restrict the definition of OOD monitoring, that tends to fuzzily embrace both OOD and OMS notions in the literature.  

\subsection{Limitations of Current Literature}

Despite an apparent convergence toward the goal of ensuring the safe operation of ML models, the fields of ODD, OOD, and OMS monitoring have largely evolved in isolation. Each community addresses a specific facet of the safety monitoring problem, remaining separated despite tackling the shared challenge of identifying and mitigating safety risks in the deployment of ML models. ODD monitors typically verify compliance with specified operational constraints but may overlook ML-specific failures that occur within these boundaries. Meanwhile, OOD and OMS methods focus on detecting anomalies in data or model behaviour but often lack integration within larger system or domain safety considerations. However, clear overlaps between the different monitoring fields can be drawn. For instance, Cheng et al. \cite{cheng2023towards} explicitly investigates how the OOD decision boundaries relate to ODD projections in the input data space, highlighting key technical challenges for aligning OOD detectors design with the operational domain. Also, Ferreira et al. \cite{ferreira2021benchmarking} presents a benchmarking framework combining OOD and OMS perspectives. Still, to date, no single monitoring paradigm comprehensively addresses the entire spectrum of ML-related threats. This gap has led to calls for hybrid approaches, as proposed in \cite{torens2025runtime} with both ODD and OOD monitoring, but there is still a lack of unified frameworks that systematically delineate the strengths and weaknesses of each approach.

There is a clear need to unify the field of ML safety monitoring by precisely defining the boundaries of its subfields to improve systematic thinking and avoid significant overlaps, and by identifying their unique potentials and complementarities. This is the goal of this paper.

\section{A Threat-Based Unification of Monitoring Approaches} \label{sec:3-my-theory}

The aim of ML safety monitors is to detect and reject incorrect predictions of the supervised model. We propose to align the various sources of ML failures with fault categories from the dependability community~\cite{avizienis2004basic}: interaction faults, development faults, and physical faults. However, at the ML level, data represents the sole point of entry for perturbation. A key challenge in applying the fault concept to ML is that most faults cannot be directly mapped onto the input space (e.g., images). While data that are outside the ODD can typically be identified as interaction faults, it is far more difficult to find inputs that will trigger development or physical faults, especially when they are in-ODD and in-distribution~\cite{guerin2023outofdistribution}. To address this, we introduce the concept of \textbf{data threats}: data points in the input space that have the potential to cause ML failures, acting as projections of certain faults onto the input domain. Yet, not all data threats produce failures, as models may be robust to challenging inputs, and some ML failures arise from faults that cannot be represented in the input space. In this section, we explore complementary monitoring approaches, starting from the previously defined concept of threats.

\subsection{Decomposition of the Input Data Space} \label{sec:3-data-domains}

\begin{figure}[t]
    \centering
    \includegraphics[width=0.98\linewidth]{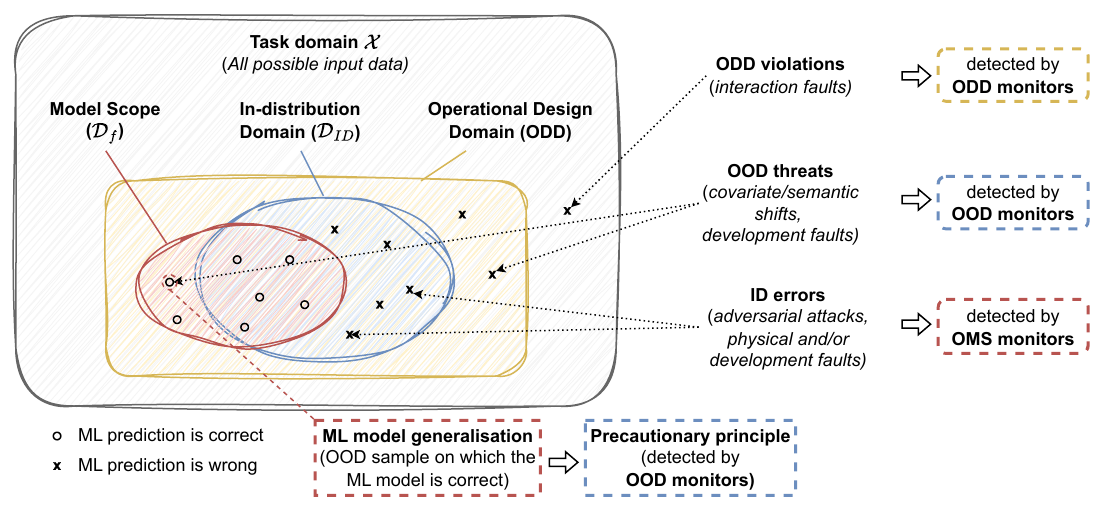}
    \vspace{-.5em}
    \caption{
        Decomposition of the ML input space and definition of data domains and data threats.
    }
    \label{fig:data-domains-threats-and-monitors}
\end{figure}

At first, we propose to project the key concepts used in the different monitoring approaches onto the ML input space (e.g., images), as depicted in the Venn diagram in Figure~\ref{fig:data-domains-threats-and-monitors} and explained hereafter. 
By projecting each monitoring concept into the input space, we highlight their mutual interactions (inclusion, exclusion, intersection) and set room for the definition of distinct categories of data threats.

First, we consider the Operational Design Domain (ODD), presented in Section~\ref{sec:2-sota-odd}, which specifies the operating conditions under which the ML system is intended and designed to function. The ODD can be defined at the system level, involving system variables (e.g., position or velocity) or environment variables (e.g., weather or lighting conditions). It is also possible to specify the ODD at the ML input level. For instance, it can be done by leveraging input-level parameters (e.g., brightness or contrast) or elements (scenery or dynamic elements) in the case of images. In the context of the VBRD task, Cappi et al.~\cite{cappi2024how} proposes to define geometric parameters (projecting system variables into the images via projection matrices) and to define domain-specific concepts identified within the images (typically used by humans to fulfil the runway detection task). For a given image, it is then possible to verify if it is compliant with the ODD, whether by checking system properties (via associated metadata) or ML input-level characteristics. We define the abstraction of the ODD in the ML input space as the set of all images that are compliant to the specified ODD, as represented in Figure~\ref{fig:data-domains-threats-and-monitors} (yellow domain). From this representation, we define our first category of data threats, \textbf{ODD violations}. They correspond to data instances that are not in the ODD, i.e., operating conditions under which the ML system is not supposed to work.

Then, we consider the datasets used to train ML models, namely $D_{train}$. A common practice in ML monitoring is to define an \textit{in-distribution} domain $\mathcal{D}_{ID}$ that contains all data instances that are drawn from the same distribution as $D_{train}$. It is also a common assumption to consider the evaluation data $D_{test}$ as drawn from the same distribution as $D_{train}$, making $\mathcal{D}_{ID}$ contain both training and evaluation data. The data points that lie inside this in-distribution domain are called ID (in-distribution) data instances, whereas the data points that are out-of $\mathcal{D}_{ID}$ are called OOD (out-of-distribution) data instances. We depict the in-distribution domain $\mathcal{D}_{ID}$ in Figure \ref{fig:data-domains-threats-and-monitors} (blue domain). Note that, in theory, the $\mathcal{D}_{ID}$ and ODD are not related to one another (depending on the data used, some OOD instances may be inside the ODD while some ID instances might be out-of the ODD). To clarify the notions, we choose to represent $\mathcal{D}_{ID}$ as included in the ODD. This choice is based on the fact that the training data is generally collected (or generated) to represent the ODD specified beforehand, as recommended by safety authorities \cite{EASA_AI_Roadmap_1_2020,EASA_AI_Concept_Paper_2_2024}. From this representation, we define a second category of data threats, \textbf{OOD threats}. They are OOD data points, i.e., outside the in-distribution domain, but still inside the ODD. Such a definition refines the existing concept of OOD data to avoid overlap with the ODD violations. Such OOD data instances depict operating conditions that should allow correct prediction. Still, potential development limitations, such as training dataset partial coverage, may lead to a more fragile ML model on these threats. We distinguish between 2 types of OOD threats: \textit{covariate} and \textit{semantic} distribution shifts threats, as described in Section \ref{sec:2-sota-ood}.

Finally, we consider the model scope $\mathcal{D}_f$, introduced in Section~\ref{sec:2-sota-oms}, which is the set of all data instances on which the ML model $f$ is correct. It is specific to the ML model at hand (two distinct ML models would have two distinct model scopes). This definition is purely theoretical as it requires to know, for any data in the ML input space, both its ground truth and the associated prediction of the ML model. Still, we define an abstraction of the model scope in the ML input space and represent it in Figure~\ref{fig:data-domains-threats-and-monitors} (red domain). We choose to represent the model scope as intersecting with the in-distribution domain ($\mathcal{D}_{ID}$). Indeed, it is most likely that the ML model performs correctly on data points similar to the ones used during training. However, as exposed in~\cite{guerin2023outofdistribution}, both domains do not completely overlap. On the one hand, some OOD data points are correctly processed by the ML model, this is called ML generalisation. On the other hand, some ID data points may result in ML failures, such situations are called in-distribution errors. We also make the choice to represent the model scope as included in the ODD. It is possible to have ML generalisation outside the ODD, nevertheless, as such out-of-ODD data should not be fed to the ML model, it is understandable to rule these cases out. In the end, from this model scope representation, we define a final category of data threats, \textbf{ID errors}, which are ID data points (inside $\mathcal{D}_{ID}$) that still provoke ML failures. As mentioned above, not all failure causes can be represented distinctly in the ML input space, making it rarely possible to understand the reasons of such ID errors. As an example, a known source of ID errors is Adversarial Attacks, malicious inputs intentionally crafted to mislead ML models while remaining imperceptible to humans~\cite{goodfellow2014explaining}.

To summarise, we projected on the ML input space the three key data domains for monitoring, that are the Operational Design Domain (ODD), the in-distribution domain ($\mathcal{D}_{ID}$), and the model scope ($\mathcal{D}_f$). This projection helps us identify the interactions between the different data domain and define specific categories of risks for the ML system.

\subsection{Categorisation of Monitoring Approaches}

For each type of data threats defined above (ODD violations, OOD data, and ID errors), we define an associated category of monitoring approaches to handle them. 

First, we define the category of \textbf{ODD monitors}. ODD monitors target inputs that are outside the projected ODD, corresponding to wrong operating conditions. Monitoring such data consists in detecting and rejecting them before they are fed to the ML model for inference. ODD monitors usually work by observing both input data and extra data from additional sensors. Indeed, it is not always possible to verify system properties from the image-level, thus requiring additional insights in the form of system variables acquired via exteroceptive or proprioceptive sensors. As they work by checking predefined rules or constraints on such data, most ODD monitors are rule-based monitors. Such monitors offer the possibility to be verified and certified. The way we define the activity of ODD monitoring is aligned with its current definition as used in the literature.

Then, we define the category of \textbf{OOD monitors}. OOD monitors aim at detecting and rejecting inputs that are out-of-distribution (OOD), i.e., data points that are not drawn from the training distribution~$\mathcal{D}_{ID}$. However, the definition we consider for OOD monitoring differs from the one commonly used. We advocate that the task of OOD monitoring should concern the detection and rejection of OOD instances, and thus, focus only on ML input data or potentially additional metadata and their distributions. On the contrary, approaches that work on internal features or outputs of ML models cannot be considered as adequate for our OOD detection paradigm. Indeed, basing the detection on signals that contain information on both the input data and the model interpretation may generate confusion: if an anomaly is detected, one cannot be sure whether it is due to an OOD data or to an ill-interpretation by the ML model itself. As mentioned in Section~\ref{sec:3-data-domains}, both \textit{covariate} and \textit{semantic} distribution shifts are OOD threats to ML models. To detect covariate distribution shifts, it is possible to work with the ML input data (images). Indeed, deriving meta-properties on the data, such as brightness or saturation for images~\cite{torens2025runtime}, can offer means to compare and draw data distributions. Such reasoning can be extended to data from additional sensors to estimate distributions on contextual variables to detect abnormal data. Most techniques used to detect covariate OOD data are data-driven as they statistically learn distributions of data properties from training data. To detect semantic distribution shifts, the task is more difficult. The key is to use complex enough methods that are able to interpret data content, especially when handling complex signals like images or videos, and to verify that it is (semantically) similar to what was seen during training. The associated monitoring approaches are also data-driven, in order to learn the underlying distribution of data points.

Finally, we define the category of \textbf{OMS monitors}. The previously described ODD and OOD monitoring activities focus on detecting threats at the input level, before the data is even processed by the model. On the opposite, the task of OMS monitoring searches for signatures of ML failures once the model has (partially or totally) processed the data. It can be viewed as seeking for actual errors (or signs of them) instead of just detecting data that is potentially dangerous. Thus, OMS monitors are designed to detect ML errors, which are provoked by internal ML faults (development or physical faults) or subtle interaction faults, such as adversarial attacks. Still, although not its primary goal, an OMS monitor can detect ``dangerous'' data instances, searched by ODD or OOD monitors, that will lead to ML failures. In practice, performing OMS translates into two considerations. First, in terms of design, any approach that monitors data produced by the ML model (either from the intermediate layers or from the output) should fall into the category of OMS monitoring. Second, in terms of evaluation, OMS monitors should always be assessed by taking into account the correctness of the ML predictions (not whether they are ID/OOD). Note that, many techniques that come from the OOD detection field as described in the literature actually fall into this class of OMS monitors. Indeed, many of them observe internal features or outputs of the ML model to detect anomalous behaviour. As mentioned above, such anomalies cannot be directly linked to OOD data threats or wrong ML model interpretations. Thus, it is more natural to consider them as out-of-model-scope (OMS) data rather than OOD data. We propose the following preliminary and non-exhaustive classification of OMS approaches, which gather techniques designed to predict ML failures.
\begin{itemize}
    \item \textbf{Feature-based methods.} 
    These approaches leverage internal features extracted from the ML model during inference. At training, the typical distribution of features extracted from correct ML predictions is captured, either statistically or by abstraction. At inference, deviations from the learnt distribution signal possible failures. Most methods labelled as ``OOD'' in the literature fall in this category, including box-abstraction monitors~\cite{he2024boxbased,henzinger2020outside}, density- and distance-based monitors~\cite{lee2018simple,sun2022outofdistribution}, outlier synthesis~\cite{du2022vos}, and introspection approaches~\cite{rahman2019did,rahman2021perframe,wilson2023safe}.

    \item \textbf{Confidence-score methods.} 
    These methods use the confidence score (and its variants) produced by the model and supposed to reflect the trust in its predictions. A score threshold, determined using training data, is used during inference to reject predictions will low confidence score. This family includes methods like Maximum Softmax Probability (MSP)~\cite{hendrycks2017baseline}, Energy scores~\cite{liu2020energybased}, ReAct~\cite{sun2021react}, or ODIN~\cite{hsu2020generalized}, mentioned in Section~\ref{sec:2-sota-ood}.

    \item \textbf{Uncertainty-based methods.} 
    These methods estimate the prediction uncertainty, without relying on the internal ML confidence score, and discard the output that are too uncertain. Such approaches include conformal prediction~\cite{andeol2023confident,zouzou2025robust}, ensemble methods~\cite{yahaya2019consensus}, and Monte Carlo dropout methods~\cite{gal2016dropout}.

    \item \textbf{Consistency-based methods.}
    These approaches check the coherence of the model predictions. For instance, in ML object detection task, some methods monitor predicted bounding box features (position, aspect ratio)~\cite{chen2021monitoring}, and others detect temporal inconsistencies, like object appearances (tickling)~\cite{kang2018model}. Such methods tend to be highly application-specific and require domain expertise.
\end{itemize}

\section{Experiment: Unified Monitoring for Vision-Based Runway Detection} \label{sec:4-my-experiment}

In this section, we illustrate the different monitoring approaches described in Section~\ref{sec:3-my-theory}, on a safety-critical application from the aviation domain, namely Vision-Based Landing (VBL). Our objective is to illustrate and validate the concepts of the diverse monitors and the benefits of using them.

\subsection{Case Study} \label{sec:41-case-study}

The VBL application, described by Cappi et al.~\cite{cappi2024how}, involves a computer vision task that consists in computing the position of the aircraft from the position of the runway within images taken during landing. We particularly focus on the Vision-Based Runway Detection (VBRD) sub-task, which involves detecting the runway in a given image, as depicted in Figure~\ref{fig:vbl-use-case}. This equates to a single-class ML object detection problem, where the objective is to accurately localise runway instances. The geometric description of the landing, including relevant positions, angles, distances, and runway markings, is illustrated in Figure~\ref{fig:vbl-use-case} (from~\cite{cappi2024how}). Specifically, the aircraft's position is defined by three parameters: the \textit{along track distance}, the \textit{vertical path angle}, and the \textit{lateral path angle}. Additionally, its attitude is defined by the Euler rotation angles: the \textit{roll} angle ($\phi$), the \textit{pitch} angle ($\theta$), and the \textit{yaw} angle ($\psi$). These six parameters, together with their defined ranges, establish a realistic aircraft trajectory for landing, referred to as the \textit{generic landing approach cone}, described in Table~\ref{tab:lard-glac-parameters}. 
\begin{table}[t]
    \centering
    \caption{
        Parameters of the generic landing approach cone.
    }
    \vspace{0.3em}
    \renewcommand{\arraystretch}{1.1}
    \begin{tabular}{|C{10em}|C{7em}||C{10em}|C{7em}|}
        \toprule
        \textbf{Parameter}   & \textbf{Range} & \textbf{Parameter}   & \textbf{Range} \\
        \midrule
        Along track distance & [0.08, 3] NM   & Roll   ($\phi$)      & [-10, 10] ° \\
        Vertical path angle  & [-2.2, -3.8] ° & Pitch  ($\theta$)    & [-8, 0] °   \\
        Lateral path angle   & [-4, 4] °      & Yaw    ($\psi$)      & [-10, 10] ° \\
        \bottomrule
    \end{tabular}
    \label{tab:lard-glac-parameters}
\end{table}
In line with Cappi et al., we define the ODD of the VBRD task of the VBL system in Definition~\ref{def:odd}, using the defined generic landing approach cone.
\begin{definition}
    \textbf{Operational Design Domain (ODD)} (VBL system, VBRD task), defined at the system level, according to the VBL intended function and the LARD data generation tool constraints:
    \begin{enumerate}
        \item The aircraft is landing on runways which present distinctive markings (piano, centreline, \dots);
        \item There exists only one runway for which current aircraft position is within the associated generic landing approach cone (Table~\ref{tab:lard-glac-parameters});
        \item The runway is fully visible on the image (no occlusion);
        \item Optimal landing conditions: clear daylight and no adverse weather conditions (fog, rain, \dots).
    \end{enumerate}
    \label{def:odd}
\end{definition}

For this case study, we use the LARD~\cite{ducoffe2023lard} dataset, which contains separate training and test sets. The training set contains 12,212 synthetic images generated using Google Earth Studio. By checking the dataset against the ODD specified in Definition~\ref{def:odd}, we find that only about 64\% (7,868/12,212) of the training images are inside the ODD\footnote[1]{We assimilate the ODD with the \textit{generic landing approach cone}, as explained in Section~\ref{sec:42-setup}.}. This illustrates the difficulties of using a unified definition of ODD, as explained in Section~\ref{sec:2-related-work}. To be consistent with the method described in Section~\ref{sec:3-my-theory}, we remove the out-of-ODD training images, to train the ML model only on data matching the expected operating conditions. Finally, we further divide our training set into a training part and a validation part, for the development of the ML component. To do so, we use a random split strategy at the runway level, i.e., by considering runways as elementary entities. Indeed, images of the same runway are too similar, and using a simple random split would result in data leak from the training set to the validation set. We eventually get a training set of 6,410 images from 22 runways and a validation set of 1,458 images from 5 runways (runways from both splits are different).

\subsection{Experimental Architecture} \label{sec:42-setup}

\begin{figure}[t]
    \centering
    \includegraphics[width=0.94\linewidth]{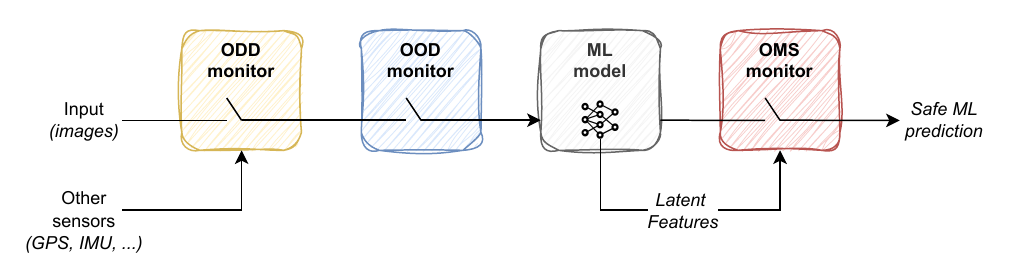}
    \vspace{-1em}
    \caption{
        Experimental setup -- serial monitoring architecture.
    }
    \label{fig:ml-monitoring-setup}
\end{figure}

The experimental setup used to illustrate the techniques of safety monitoring is depicted in Figure~\ref{fig:ml-monitoring-setup} and further described in this subsection. We adopt a serial architecture, which naturally follows from the characteristics of the three proposed monitors. 
Indeed, ODD and OOD monitoring do not require running the ML model and should therefore be applied prior to prediction, whereas OMS requires analysis of the model output and must be applied after prediction. Additionally, since in-distribution data are a subset of ODD data, it is logical to apply ODD filtering first. The complete source code is available on a public repository\footnote[2]{\url{https://github.com/mathieudario/swmf-lard-yolo-ICPR26}}. We used Python 3.10 and Pytorch to run our experiments.

\subsubsection{ML Algorithm.}
For our study, we use an object detection model to detect runway instances in images. As the application lies in the context of real-time embedded safety-critical systems, we decide to implement a single-stage object detector, that suits the real-time inference context. Thus, we use a pretrained YOLOv5-n, a reduced version of YOLOv5 (You Only Look Once) models proposed by Ultralytics, and tune it for 20 epochs on our training and validation sets.

\subsubsection{ODD Monitor.}
In this part, we implement an ODD monitor (yellow module, Figure~\ref{fig:ml-monitoring-setup}) to verify the compliance of the input data with the specified operational conditions. We suppose that we have access to data from additional sensors during operations. For instance, for an image sent to the ML model, we should have access to the associated aircraft's position and attitude. The aircraft's position could be obtained using GPS-like sensors, while its attitude could be obtained using IMU-like sensors. We then use the ODD of the VBRD task (Definition~\ref{def:odd}) and restrict the monitoring to checking compliance with the landing cone (Table~\ref{tab:lard-glac-parameters}). This is due to a limited amount of available metadata for each given image. Indeed, in the LARD dataset, the available metadata are the airport ID, the runway ID, the aircraft's position (\textit{along track distance}, \textit{vertical path angle}, \textit{lateral path angle}) and attitude (\textit{roll}, \textit{pitch}, \textit{yaw}), as well as the estimated time-of-day. However, no information of weather or lighting conditions is available, given the inherent constraints of the synthetic data generator~\cite{cappi2024how}.

\subsubsection{OOD Monitor.}
To monitor the presence of OOD data, we implement an OOD detector (blue module, Figure~\ref{fig:ml-monitoring-setup}), inspired by Torens et al.~\cite{torens2025runtime}. As images make direct distribution estimation challenging, we extract four meta-properties from each image: \textit{brightness}, \textit{saturation}, \textit{entropy}, and \textit{edge-amount}, which together capture basic aspects of image appearance. In line with~\cite{torens2025runtime}, we compute these properties for all training images and model their distributions with beta distributions. A quantile threshold (e.g., 1\%) is then determined for each property, identifying their most extreme values, i.e., regions where the model has seen little or no training data and, thus, could not be trusted. At runtime, we compute the same four properties from any input image and check if their values fall within the learnt quantile intervals. If any property is outside its corresponding range, the image is flagged as OOD. This approach provides a straightforward, one-dimensional method for detecting OOD data. While more complex multi-dimensional methods exist, we limit our discussion here to this simple technique to clearly illustrate the principle of OOD monitoring.

\subsubsection{OMS Monitor.}
To monitor the presence of OMS data, we use a box-abstraction monitor (red module, Figure~\ref{fig:ml-monitoring-setup}), inspired by previous work on YOLO models \cite{he2024boxbased}. This OMS monitor works on internal features (vector(s) of values of model intermediate layer) extracted from the ML model at runtime. This approach is a post-hoc method that does not necessitate to retrain the ML model, which aligns with the gray-box monitoring paradigm (SMs can observe any input, state, or output of the ML model but cannot modify it, e.g., retrain it). As mentioned in~\cite{henzinger2020outside}, box-abstraction monitors are fast and are based on a rather simple algorithm to decide whether an extracted feature is in- or out-of-model-scope. The training procedure of the box-based monitor is taken from He et al.~\cite{he2024boxbased}. First, for each image in the training set, we extract internal features from the logits layer, i.e., the last layer before the model output, corresponding to each detected bounding box in the model prediction. Then, applying the training trick introduced by~\cite{guerin2023outofdistribution}, we only keep the logits corresponding to correct detections of the model (True Positives). Finally, we partition the extracted logits $F$ into a set of $k$ subsets, denoted as $\pi(F) = \{F^1, \dots, F^k \}$. For each subset $F^j, \;j \in [1, \dots, k]$, we build a \textit{tight-box-abstraction} $B^j$ (as in~\cite{he2024boxbased}), resulting in a global box-abstraction $\mathcal{B} = \{ B^1, \dots, B^k \}$. At runtime, for a new image, we extract features from the logits layer. The OMS monitor will reject a detection $p_i = (bbox_i, label_i, conf_i)$ if the corresponding logits vector $z_i$ is not contained in $\mathcal{B}$, i.e., $\not\exists j \in [1, \dots, k], \; z_i \in B^j$. 

\subsection{Evaluation Data} \label{sec:43-corrupt-dataset}

\begin{figure}[t]
    \centering
    \begin{subfigure}{0.195\textwidth} 
        \centering
        \includegraphics[width=0.99\linewidth]{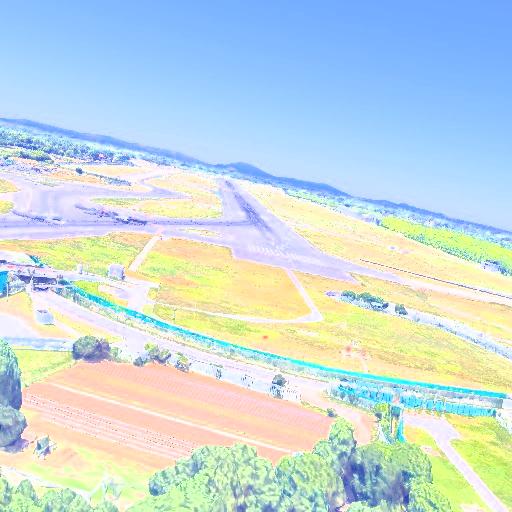}
        \caption{Brightness}
        \label{fig:im-corrupt-brightness}
    \end{subfigure}
    \hfill
    \begin{subfigure}{0.195\textwidth} 
        \centering
        \includegraphics[width=0.99\linewidth]{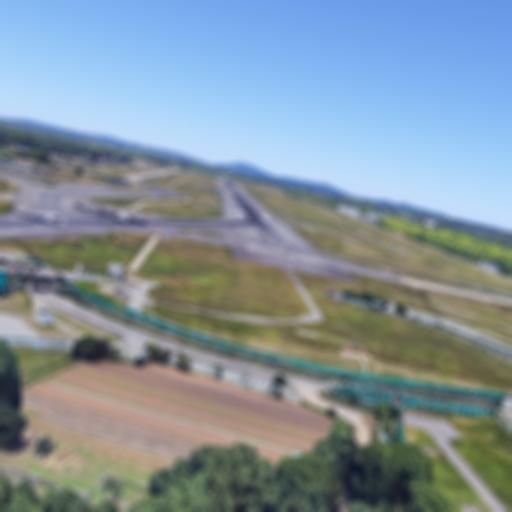}
        \caption{Defocus blur}
        \label{fig:im-corrupt-defocus-blur}
    \end{subfigure}
    \hfill
    \begin{subfigure}{0.195\textwidth} 
        \centering
        \includegraphics[width=0.99\linewidth]{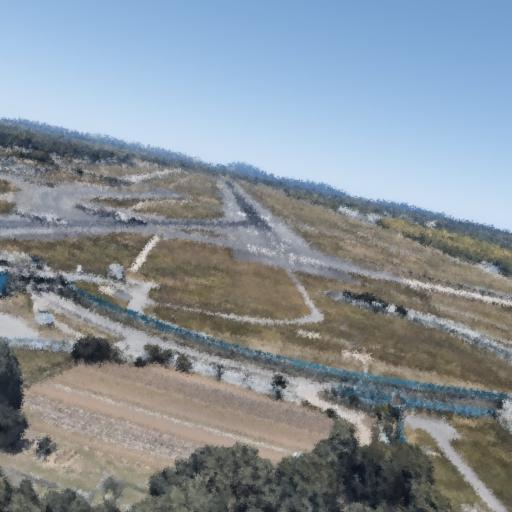}
        \caption{Frosted blur}
        \label{fig:im-corrupt-frosted-blur}
    \end{subfigure}
    \hfill
    \begin{subfigure}{0.195\textwidth} 
        \centering
        \includegraphics[width=0.99\linewidth]{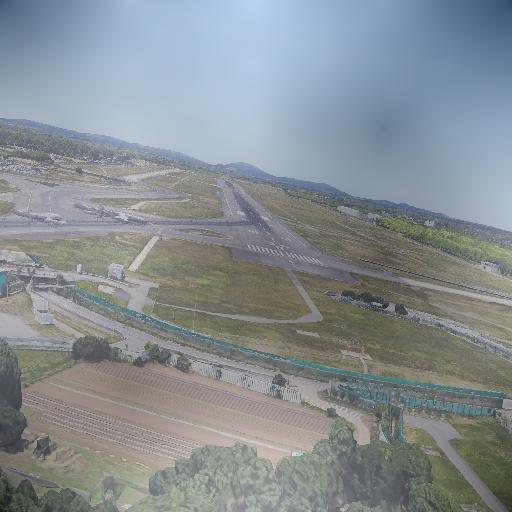}
        \caption{Fog}
        \label{fig:im-corrupt-fog}
    \end{subfigure}
    \hfill
    \begin{subfigure}{0.195\textwidth} 
        \centering
        \includegraphics[width=0.99\linewidth]{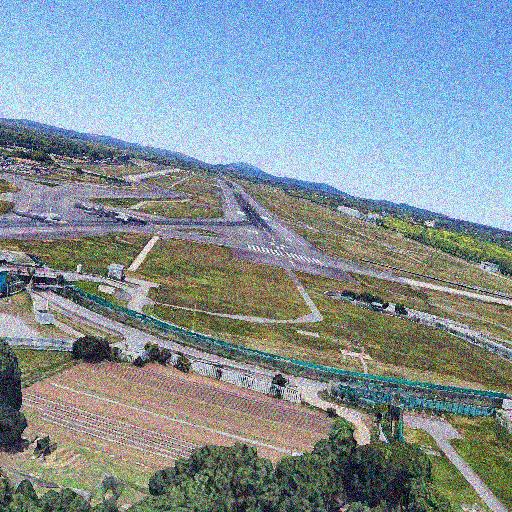}
        \caption{Gaussian noise}
        \label{fig:im-corrupt-gaussian-noise}
    \end{subfigure}%
    \vspace{-.5em}
    \caption{
        Corruptions applied to the image in Figure~\ref{fig:vbl-use-case}, for the highest severity level.
    }
    \label{fig:lard-corruptions}
\end{figure}

We conduct our evaluation using two separate test sets, resulting in two distinct experiments. The first test set is sourced from the LARD~\cite{ducoffe2023lard} dataset and includes both synthetic and real footage images. For this experiment, we only keep the 2,212 synthetic images, to avoid having to do additional domain adaptation, as the model is trained on synthetic images only. These images correspond to 79 unique runways, all different from those used in training and validation (see Section~\ref{sec:41-case-study}). Notably, this nominal test set contains out-of-ODD, out-of-distribution, and out-of-model-scope images (see Section~\ref{sec:45-results}). From this nominal test set, we create a set of corrupted images specifically designed to evaluate the OOD and OMS monitors. These degradations affect only the images, leaving the associated metadata (aircraft's position and attitude) unchanged and making the corrupted data undetectable by ODD monitoring. We apply five types of corruptions from Hendrycks and Dietterich~\cite{hendrycks2018benchmarking}: \textit{brightness}, \textit{defocus blur}, \textit{frosted blur}, \textit{fog}, and \textit{Gaussian noise}. These corruptions represent both external factors (e.g., weather, lighting conditions) and hardware issues (e.g., noise, calibration defects). Figure~\ref{fig:lard-corruptions} illustrates these corruptions at their highest severity level to highlight the effect of each degradation. In total, we implement three severity levels (small, medium, large) for each corruption, resulting in 15 different corruptions and 15 $\times$ 2,212 images.

\subsection{Evaluation Metrics}

To evaluate monitors, we first need to assess the performance of the underlying supervised model. In particular, we need to know when the model is correct or not, to derive the benefits of using monitors.

\subsubsection{Metrics for Object Detection.}
To evaluate an object detector, one commonly computes metrics like the precision $P$ and the recall $R$. In this context, a True Positive ($TP$) is when a bounding box predicted by the model is similar enough to a bounding box of a ground truth object, i.e., we say that a detection is matched to a ground truth. On the contrary, a False Positive ($FP$) corresponds to a predicted bounding box that matches no ground truth object. Finally, a False Negative ($FN$), is a ground truth bounding box that have not been matched with any predicted box. One of the metrics used to evaluate the similarity between two bounding boxes $A$ and $B$ is the \textit{Intersection-over-Union} (IoU) defined as $IoU = \frac{A \cap B}{A \cup B}$. Hence, an IoU-threshold has to be determined beforehand to decide whether two bounding boxes are similar or not. In most applications, an IoU-threshold of $0.5$ is used. However, in safety-critical applications, higher thresholds can be used, reflecting the need for a higher localisation accuracy of the ML model. For our safety-critical case, we use $\tau_{IoU} = 0.7$, evaluating the ML model in a rather strict way. Finally, a prediction of an object detector is often composed of three elements: a bounding box (spatial localisation in an image), a class label associated to the object inside the bounding box, and a confidence score reflecting the model certainty in its detection. For a deployed model in inference, one usually has to set a confidence threshold $\tau_{conf}$ to only output the ML detections that have a confidence score above this threshold. Changing the confidence threshold influences the number of ML detections, and thus, the number of $TP$, $FP$, and $FN$. Such confidence threshold is commonly computed to optimize some performance metric, like the $F_1$-score, which offers a compromise between a minimal ratio of wrong detections ($P$) and a maximal ratio of detected ground truths ($R$). For our application, we compute $\tau_{conf}^{[\tau_{IoU}=0.7]} = 0.695$.

\subsubsection{Metrics for Safety Monitors.}
To evaluate safety monitors, we use the safety-oriented metrics introduced by Guerin et al.~\cite{guerin2022unifying}. For each monitor $m_f$, we evaluate its safety benefits through the Safety Gain ($SG$), the remaining safety gaps after using it via the Residual Hazard ($RH$), and the negative impact on the monitored system via the Availability Cost ($AC$). In our specific application, since inputs are considered one-by-one in an independent manner, we simplify the return functions of~\cite{guerin2022unifying} as they only depend on the model's and monitors' actions on individual images. Also, for a given input $x$, if the associated model prediction $f(x)$ contains either a false positive ($FP$) or false negative ($FN$), then we consider that the model is wrong on the whole input $x$. With such considerations, we re-define the Safety Gain (Equation~\ref{eq:sg-relax}), as the proportion of dangerous situations (images) correctly filtered by the monitor; the Residual Hazard (Equation~\ref{eq:rh-relax}), as the proportion of dangerous images missed (accepted by the monitor); and the Availability Cost (Equation~\ref{eq:ac-relax}), as the proportion of safe situations wrongly rejected by the monitor. Safety metrics are usually estimated on an evaluation dataset ($D=\{ (x_i,y_i) \}^n_{i=1}$). As shown in~\cite{guerin2022unifying}, when monitoring is viewed as a binary classification (0~=~accept, 1~=~reject) task, these metrics are related to those used in classifiers evaluation, such as the monitor's True Positives ($TP^{[m_f]}$), False Positives ($FP^{[m_f]}$), and False Negatives ($FN^{[m_f]}$).
\begin{equation}
SG_{m_f}^{D} 
    = \frac{1}{n} \sum^{n}_{i=1} \left( \begin{cases}
        1 & \text{if } f(x_i) \neq y_i ,\; m_f(x_i) = 1 \\
        0 & \text{otherwise}
    \end{cases} \right)
    = \frac{1}{n} TP^{[m_f]}
    \label{eq:sg-relax}
\end{equation}
\begin{equation} 
RH_{m_f}^{D} 
    = \frac{1}{n} \sum^{n}_{i=1} \left( \begin{cases}
        1 & \text{if } f(x_i) \neq y_i ,\; m_f(x_i) = 0 \\
        0 & \text{otherwise}
    \end{cases} \right) 
    = \frac{1}{n} FN^{[m_f]}
    \label{eq:rh-relax}
\end{equation}
\begin{equation}
AC_{m_f}^{D}
    = \frac{1}{n} \sum^{n}_{i=1} \left( \begin{cases}
        1 & \text{if } f(x_i) = y_i ,\; m_f(x_i) = 1 \\
        0 & \text{otherwise}
    \end{cases} \right) = \frac{1}{n} FP^{[m_f]}
    \label{eq:ac-relax}
\end{equation}

\subsection{Results} \label{sec:45-results}

\subsubsection{Experiment 1.}

\begin{figure}[t]
    \centering
    \begin{subfigure}{0.6\textwidth}
        \centering
        \includegraphics[width=\linewidth]{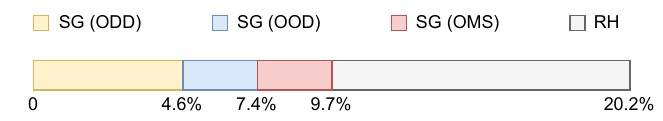}
    \end{subfigure}\\%
    \vspace{1.0em}
    \begin{subfigure}{0.6\textwidth}
        \centering
        \includegraphics[width=\linewidth]{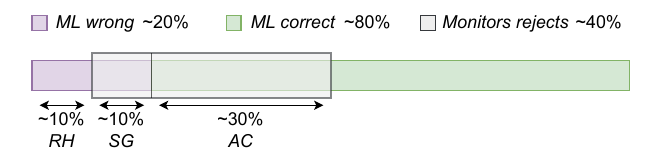}
    \end{subfigure}
    \vspace{-.5em}
    \caption{
        Visualisation of the cumulated Safety Gain ($SG$) and Residual Hazard ($RH$) (\textit{top}), and the proportions of monitors rejections over all images when using (ODD+OOD+OMS) monitors (\textit{bottom}).
    }
    \label{fig:cumul-metrics}
    \vspace{-.5em}
\end{figure}

First, we evaluate the monitors using the nominal test set of 2,212 images, each containing exactly one ground truth object (see Section~\ref{sec:43-corrupt-dataset}). The YOLOv5-n model achieves a Precision $P=0.869$ and a Recall $R=0.781$ on this set, resulting in a Residual Hazard of 0.202, meaning that roughly 20\% of images induce errors in the model (Table~\ref{tab:safety-metrics-test}, first row, left). Next, we observe that combining different types of monitors leads to nearly match the sum of the gains from each individual monitor, demonstrating the complementarity of our proposed monitoring categories (see Section~\ref{sec:3-my-theory}). This effect is further highlighted in Figure~\ref{fig:cumul-metrics} (\textit{top}), which depicts the contribution of each monitor to the safety gain, when all monitors (ODD, OOD, and OMS) are sequentially applied. However, this complementarity comes at the expense of an increased Availability Cost ($AC$). As more monitors are applied, the cost to system availability rises. As shown in Figure~\ref{fig:cumul-metrics} (\textit{bottom}), deploying all monitors leads to the correct rejection of about 50\% of ML error sources, but for an availability cost of about $0.3$. This illustrates the principal challenge of monitoring: substantial safety improvements are often offset by a significant rate of false alarms ($FP^{[m_f]}$). 
\begin{table}[t]
    \centering
    \caption{
        Results of experiment 1, performances of monitors on the nominal test set.
    }
    \vspace{0.2em}
    \label{tab:safety-metrics-test}
    \renewcommand{\arraystretch}{1.1}
    \hfill
    \begin{subtable}[c]{0.44\textwidth}
        \centering
        \begin{tabularx}{\linewidth}{C{9em}Y*{3}{Y}}
            \toprule
            \textbf{Monitors} & 
            \textbf{SG} $\uparrow$ & \textbf{RH} $\downarrow$ & \textbf{AC} $\downarrow$ \\
            \midrule
            No monitor  & 0.0 & 0.202 & 0.0 \\
            ODD         & 0.046 & 0.156 & 0.159 \\
            OOD         & 0.041 & 0.161 & 0.150 \\
            OMS         & 0.040 & 0.162 & 0.066 \\
            \bottomrule
        \end{tabularx}
    \end{subtable}%
    \hfill
    \begin{subtable}[c]{0.44\textwidth}
        \centering
        \begin{tabularx}{\linewidth}{C{9em}Y*{3}{Y}}
            \toprule
            \textbf{Monitors} & 
            \textbf{SG} $\uparrow$ & \textbf{RH} $\downarrow$ & \textbf{AC} $\downarrow$ \\
            \midrule
            ODD+OOD     & 0.074 & 0.127 & 0.275 \\
            ODD+OMS     & 0.074 & 0.127 & 0.203 \\
            OOD+OMS     & 0.071 & 0.131 & 0.202 \\
            ODD+OOD+OMS & 0.097 & 0.105 & 0.311 \\
            \bottomrule
        \end{tabularx}
    \end{subtable}    
    \hfill
\end{table}

\subsubsection{Experiment 2.}

\begin{table}[t]
    \vspace{-1.em}
    \centering
    \caption{
        Results of experiment 2, performances on data with corruption \textbf{C} and severity level \textbf{S}.
    }
    \vspace{0.2em}
    \label{tab:safety-metrics-test-corrupt}
    \begin{tabularx}{1.\textwidth}{L{6.5em}C{2em}C{6.5em}YYYYYYYYY}
    \toprule
     &  & \multicolumn{1}{c}{\textbf{No monitor}} & \multicolumn{3}{c}{\textbf{OOD}} & \multicolumn{3}{c}{\textbf{OMS}} & \multicolumn{3}{c}{\textbf{OOD+OMS}} \\
    \cmidrule(lr){3-3} \cmidrule(lr){4-6} \cmidrule(lr){7-9} \cmidrule(lr){10-12}
    \textbf{C} & \textbf{S} & \textbf{RH} $\downarrow$ &
    \textbf{SG} $\uparrow$ & \textbf{RH} $\downarrow$ & \textbf{AC} $\downarrow$ & 
    \textbf{SG} $\uparrow$ & \textbf{RH} $\downarrow$ & \textbf{AC} $\downarrow$ & 
    \textbf{SG} $\uparrow$ & \textbf{RH} $\downarrow$ & \textbf{AC} $\downarrow$ \\
    \midrule
    \multirow{3}{*}{Brightness}
     & 1 & 0.281 & 0.238 & 0.043 & 0.572 & 0.066 & 0.215 & 0.085 & 0.248 & 0.033 & 0.589 \\
     & 2 & 0.451 & 0.446 & 0.005 & 0.537 & 0.128 & 0.323 & 0.090 & 0.447 & 0.005 & 0.539 \\
     & 3 & 0.605 & 0.605 & 0.000 & 0.395 & 0.204 & 0.401 & 0.098 & 0.605 & 0.000 & 0.395 \\
    \addlinespace
    \multirow{3}{*}{Defocus Blur}
     & 1 & 0.246 & 0.073 & 0.173 & 0.267 & 0.055 & 0.192 & 0.082 & 0.105 & 0.141 & 0.317 \\
     & 2 & 0.421 & 0.421 & 0.000 & 0.579 & 0.094 & 0.327 & 0.116 & 0.421 & 0.000 & 0.579 \\
     & 3 & 0.616 & 0.616 & 0.000 & 0.384 & 0.130 & 0.486 & 0.154 & 0.616 & 0.000 & 0.384 \\
    \addlinespace
    \multirow{3}{*}{Fog}
     & 1 & 0.242 & 0.089 & 0.153 & 0.311 & 0.050 & 0.192 & 0.076 & 0.115 & 0.126 & 0.351 \\
     & 2 & 0.261 & 0.199 & 0.062 & 0.587 & 0.057 & 0.204 & 0.086 & 0.211 & 0.050 & 0.602 \\
     & 3 & 0.293 & 0.281 & 0.012 & 0.666 & 0.080 & 0.213 & 0.090 & 0.284 & 0.009 & 0.672 \\
    \addlinespace
    \multirow{3}{*}{Frosted Blur}
     & 1 & 0.272 & 0.052 & 0.220 & 0.144 & 0.049 & 0.223 & 0.092 & 0.089 & 0.184 & 0.218 \\
     & 2 & 0.386 & 0.060 & 0.327 & 0.134 & 0.062 & 0.326 & 0.108 & 0.112 & 0.275 & 0.214 \\
     & 3 & 0.407 & 0.284 & 0.124 & 0.433 & 0.083 & 0.324 & 0.104 & 0.306 & 0.101 & 0.460 \\
    \addlinespace
    \multirow{3}{*}{Gaussian Noise}
     & 1 & 0.319 & 0.319 & 0.000 & 0.679 & 0.039 & 0.280 & 0.114 & 0.319 & 0.000 & 0.679 \\
     & 2 & 0.600 & 0.600 & 0.000 & 0.400 & 0.037 & 0.563 & 0.146 & 0.600 & 0.000 & 0.400 \\
     & 3 & 0.816 & 0.816 & 0.000 & 0.184 & 0.047 & 0.769 & 0.085 & 0.816 & 0.000 & 0.184 \\
    \bottomrule
    \end{tabularx}
    \vspace{-1.em}
\end{table}

Next, we evaluate the OOD and OMS monitors on the corrupted test set (see Section~\ref{sec:43-corrupt-dataset}), with results presented in Table~\ref{tab:safety-metrics-test-corrupt}. Overall, the YOLOv5-n model performs worse on this set compared to the nominal test set. This decline is visible in the higher global hazard ($SG + RH$), which exceeds 0.202 (global hazard from the first experiment) for all corruption types and severity levels.

First of all, we observe that the OOD monitor is highly active across most corruptions and severity levels, resulting in strong Safety Gains ($SG$) and low Residual Hazards ($RH$). This highlights the OOD monitor’s effectiveness in rejecting corrupted images, a task for which it is specifically designed for (see Section~\ref{sec:3-my-theory}). However, since the OOD monitor makes decisions independently of the ML model’s robustness, it may reject images even when the model performs well. Following such precautionary principle (see Figure~\ref{fig:data-domains-threats-and-monitors}) can result in substantial Availability Costs ($AC$) for the OOD monitor, e.g., for \textit{Fog} or \textit{Gaussian Noise} corruptions. Nonetheless, given that these corruptions are relatively rare in practice, a high rejection rate for “abnormal” data is acceptable.

In contrast, the OMS monitor is less effective at detecting corrupted instances that cause ML errors. It globally achieves lower Safety Gains ($SG$) and Availability Costs ($AC$), indicating less frequent activation, and when considering both OOD and OMS, the action of the OMS monitor seems negligible compared to the OOD's. This clearly contrasts with its strong performance in the first experiment, where it provided similar Safety Gains at much lower Availability Costs compared to other monitors. This shows the limitations of OMS monitors when addressing corrupted (OOD) data, emphasizing the advantage of employing a sequential monitoring setup. Because OMS monitors rely on complex mechanisms tied to the supervised model, they are more sensitive to perturbations that affect this model. Thus, it is logical to first filter out clear OOD data using OOD monitors, and then use the OMS monitor to make more nuanced decisions within the remaining ID domain.

\subsubsection{Qualitative Analysis.}

Beyond the quantitative results, our framework offers significant qualitative advantages. Firstly, it enables comprehensive evaluation by assessing all monitors with consistent metrics ($SG$, $RH$, and $AC$), facilitating straightforward comparison of the strengths and limitations of each monitoring approach and making it easier to identify complementarities. Secondly, the framework simplifies the design of monitoring setups by explicitly linking each monitoring strategy to its targeted threat, providing clarity and direction during integration. Furthermore, the modular nature of the framework supports flexible adaptation to diverse application domains and evolving requirements.
\vspace{-0.5em}

\section{Conclusion and Perspectives} \label{sec:5-conclusion}

In this paper, we proposed a unified framework for safety monitoring of ML-based systems, aiming to bridge monitoring approaches originating from distinct research communities, notably machine learning and system safety. The proposed framework structures existing methods into three complementary categories: Operational Design Domain (ODD) monitoring, Out-of-Distribution (OOD) monitoring, and Out of Model Scope (OMS) monitoring. While this categorization is inspired by the existing literature, a key contribution of this work lies in the clarification of the underlying concepts associated with each category, as well as in the explicit analysis of their overlaps and complementarities.

The separation of monitoring approaches into three distinct categories proved beneficial both conceptually and practically. It enabled the analysis and implementation of monitoring techniques without conceptual overlap, thereby simplifying their integration within a single system architecture. Experimental results confirm that ODD, OOD, and OMS monitors address different classes of threats, and are therefore inherently complementary. However, the results also highlight that the current levels of system availability achieved by inclusion of individual monitors remain insufficient for deployment in safety-critical systems. Although the quantitative analysis of availability costs was beyond the scope of this paper, it clearly emerges as a critical limitation. Importantly, the proposed unification also facilitates a unified evaluation methodology, which contributes to the justification of trustworthiness when combining multiple monitors within a single safety architecture.

Several perspectives arise from this work. First, the proposed framework will be fully validated only when high performance levels are achieved for each type of monitor. In this context, the availability cost associated with safety monitoring remains a major open challenge shared by ODD, OOD, and OMS approaches. Second, the evaluation and justification of combined monitoring strategies should rely on a broader set of metrics and analytical tools than those considered in this paper, in order to better capture safety and availability trade-offs. Finally, data selection plays a central role in the relevance and effectiveness of monitoring approaches. Since monitoring training and evaluation strongly depends on the representativeness of data with respect to targeted threats, further research is needed to define principled methods for data collection and curation aligned with operational contexts. 

\subsubsection*{Acknowledgment.}
This work has benefited from the AI cluster ANITI2 funded by the French government through the ANR under the France 2030 program (grant ANR-23-IACL-0002).

\bibliographystyle{splncs04}
\bibliography{references}

\end{document}